\documentclass[10pt,twocolumn,letterpaper]{article}

\usepackage{cvpr}
\usepackage{times}
\usepackage{epsfig}
\usepackage{graphicx}
\usepackage{amsmath}
\usepackage{amssymb}
\usepackage{comment}
\usepackage{enumitem}
\usepackage{balance}

\begin{document}


\author{
    Kaniz Fatima\textsuperscript{1,5},
    Michael Schuckers\textsuperscript{2,6},
    Gerardo Cruz-Ortiz\textsuperscript{3,7},
    Daqing Hou\textsuperscript{1,8},\\
    Sandip Purnapatra\textsuperscript{1,9},
    Tiffany Andrews\textsuperscript{3,10},
    Ambuj Neupane\textsuperscript{3,11},\\
    Brandeis Marshall\textsuperscript{4,12},
    Stephanie Schuckers\textsuperscript{1,13}\\
    {\normalsize \textsuperscript{1}Electrical and Computer Engineering, Clarkson University, Potsdam, NY, USA}\\
    {\normalsize \textsuperscript{2}School of Data Science, Univ. of North Carolina Charlotte, Charlotte, NC, USA}\\
    {\normalsize \textsuperscript{3}General Services Administration (GSA), USA}    
    {\normalsize \textsuperscript{4}DataedX Group LLC, USA}\\
    {\normalsize  \texttt{\textsuperscript{5}fatimak@clarkson.edu,\textsuperscript{6}schuckers@charlotte.edu,\textsuperscript{7}gerardo.cruz-ortiz@gsa.gov,}}\\
    {\normalsize \texttt{\textsuperscript{8}dhou@clarkson.edu,\textsuperscript{9}purnaps@clarkson.edu,\textsuperscript{10}tiffanyj.andrews@gsa.gov,}}\\
    {\normalsize \texttt{\textsuperscript{11}ambuj.neupane@gsa.gov,\textsuperscript{12}brandeis@dataedx.com,\textsuperscript{13}sschucke@clarkson.edu}}\\    
}
\title{A large-scale study of performance and equity of commercial remote identity verification technologies across demographics}

\maketitle
\thispagestyle{empty}

\begin{abstract}
\vspace*{-0.1in}
As more types of transactions move online, there is an increasing need to verify someone's identity remotely. Remote identity verification (RIdV) technologies have emerged to fill this need. RIdV solutions typically use a smart device to validate an identity document like a driver's license by comparing a face selfie to the face photo on the document. Recent research has been focused on ensuring that biometric systems work fairly across demographic groups. This study assesses five commercial RIdV solutions for equity across age, gender, race/ethnicity, and skin tone across 3,991 test subjects. This paper employs statistical methods to discern whether the RIdV result across demographic groups is statistically distinguishable.  Two of the RIdV solutions were equitable across all demographics, while two RIdV solutions had at least one demographic that was inequitable. For example, the results for one technology had a false negative rate of 10.5\% +/-
4.5\% and its performance for each demographic category was within the error bounds, and, hence, were equitable. The other technologies saw either poor overall performance or inequitable performance. For one of these, participants of the race Black/African American (B/AA) as well as those with darker skin tones (Monk scale 7/8/9/10) experienced higher false rejections.  Finally, one technology demonstrated more favorable but inequitable performance for the Asian American and Pacific Islander (AAPI) demographic.  This study confirms that it is necessary to evaluate products across demographic groups to fully understand the performance of remote identity verification technologies. 

\end{abstract}
\vspace*{-0.1in}

\section{Introduction}


Remote identity verification (RIdV) is necessary as individuals seek to perform more sensitive transactions online rather than in-person, such as opening a bank account, accessing government services, crossing a border, and applying for a loan.  Some RIdV technologies use a combination of document verification and a one-to-one (1:1) biometric comparison between the photo on the document and a selfie. It is essential to ensure these AI-based solutions treat individuals fairly, regardless of variables such as race, gender, age, or any other demographic factors. This issue has gained significant attention due to the observed biases within some face recognition systems, which can lead to errors and discriminatory outcomes, especially affecting marginalized groups. A notable study by Buolamwini et al. \cite{gender-shades-1} brought attention to a higher error rate in gender classification among darker-skinned females compared to other groups. Although the focus was on gender classification, the study underscored broader concerns about fairness in face recognition technology. 

\begin{figure}[!t]
\centering


\begin{minipage}[htbp]{0.5\textwidth}
    \includegraphics[width=\textwidth]{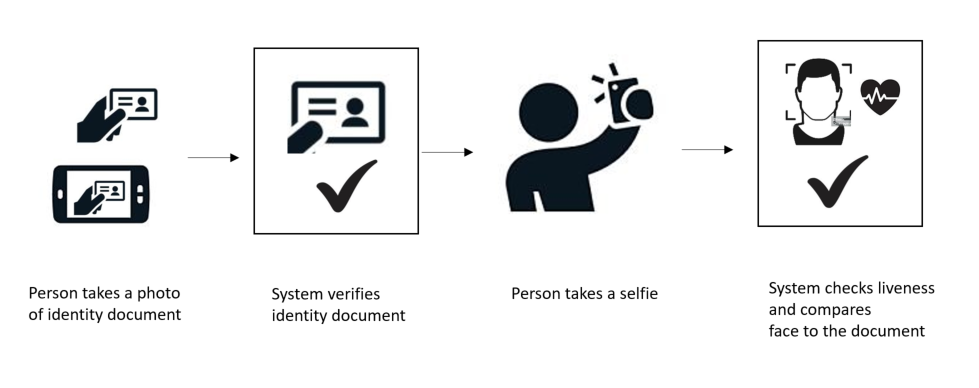}
    \label{fig:race}
    \end{minipage}
\hfill

\caption{Typical process by which a user interacts with remote identity verification technologies  on their own devices via a webpage or a software app}
\label{rIDV_process}

\end{figure}

This study focuses on remote identity verification software and investigates statistical equity across diverse demographic groups, including age, gender, race/ethnicity, and skin tone. Prior work by NIST and others~\cite{grother2019face, NIST-FRVT,Demographic-Effects} have considered the fairness of face matching systems. 
 Our work expands upon prior work by testing full end-to-end remote identity verification systems which include the face matcher, as well as the user interface, capture process, document verification check, and liveness check.  The end-to-end process is shown in Figure \ref{rIDV_process}.

\section{Background}



Howard et al. \cite{howard2019effect}, as part of a study of equity in face recognition, proposed the term “differential performance” to describe biometric performance differences across demographic groups. Previous studies have demonstrated that (some) biometric recognition technologies have shown differential performance for some demographic groups
\cite{Demographic-Effects, skin-tone,cavazos2020accuracy, li2021information, drozdowski2020demographic, howard2019effect, karkkainen2021fairface, robinson2020face, sukthanker2022importance, yucer2022measuring, chouldechova2022unsupervised}.
 A survey of the literature reveals that demographic factors have a significant influence on the performance of facial biometric algorithms and, for some algorithms, there is a poorer biometric performance for females, dark-skinned females, and the youngest subjects although which demographic is impacted varies across algorithms,  and, in some cases, do not show evidence of differential performance. 

Research has also considered ways to mitigate these problems through creation of more diverse training sets and inclusive training strategies,  e.g., \cite{wang2019racial, gong2020jointly, gong2021mitigating}.  Bias related to face presentation attack detection (PAD) algorithms have also been studied \cite{alshareef2021study, singh2020robustness}.

Most research studies have focused on the matching algorithm based on an existing face image dataset. According to the International Organization for Standardization (ISO) standard 19795-1 \cite{ISO_IEC_19795-1:2021}, this type of evaluation would be considered a “technology evaluation”, defined as “offline valuation of one or more algorithms for the same biometric modality using a pre-existing or especially-collected corpus of samples'' \cite{ISO_IEC_19795-1:2021}. ISO defines a scenario evaluation as an “evaluation that measures end-to-end system performance in a prototype or simulated application with a test crew”. A scenario evaluation adds the full context of the use case and includes the end-to-end system which incorporates image-capture hardware and software, user experience, quality control, etc. Testing the algorithms in a technology evaluation is an important initial phase for testing a biometric recognition system. However, it should be complemented by practical testing of the end-to-end system in real-life situations. This ensures that the effect of all components is fully understood for the system's intended real-world application. This study is a scenario evaluation of remote identity verification and includes an assessment of the facial matching algorithm as well as the user interface, document verification, and presentation attack detection (PAD).  For this study, the false negative rate (FNR) will be used to assess the system, where an error is defined as the case where a genuine person who matches their document is falsely rejected.  A rejection could be due to any component in the software (capture, PAD, matcher, etc).

National Institute of Standards and Technology’s (NIST) Face Recognition Vendor Test (FRVT) is one of the most extensive technology evaluations in the world \cite{grother2019face, NIST-FRVT}.  Commercial software biometric algorithms are submitted to NIST for testing.  Evaluation is performed across a variety of datasets including border, visa application, and mugshot images, and for both identification (1:N) and verification (1:1).  Performance is reported in terms of false non-match rate (FNMR) and false match rate (FMR) for verification and false negative identification rate (FNIR) and false positive identification rate (FPIR) for identification.  
FRVT results consider demographics including country of origin, age, and gender, and confidence intervals are provided based on bootstrapping.  Results are continually updated at \cite{NIST-FRVT}.  FRVT is only focused on the facial recognition algorithm, not the full end-to-end system.  FRVT does not consider the user interface, document verification, or presentation attack detection (PAD) steps.

Recently, the Department of Homeland Security Biometric Technology Rallies announced a Remote Identity Validation Technology Demonstration \cite{rivtd2023}.  The rally separates the components of remote identity verification into three parts (document validation, match to document, and face liveness) and has a demonstration for each part.  The study in our paper has similarities to the DHS’s Rally with important distinctions.  The study presented herein is conducted in a full scenario evaluation with all components integrated.  Additionally, the evaluation is performed in the subject’s own environment and thus imaging conditions are more diverse and impacted by factors such as camera model, environmental lighting and setting, participant’s proficiency in taking pictures, etc. 

The need for this work is further underscored by NIST’s request for information related to establishing “metrics and testing methodologies to allow for assessment of performance and understanding of impacts across user populations (e.g., bias in artificial intelligence)” and recommending “operational testing to determine if the image capture technologies have introduced unintentional biases” in the draft Enrollment and Identity Proofing (SP-800-63A-4) document \cite{DigitalIdentityGuidelines}. 

In 2019, Federal Agencies were mandated to use NIST's 800-63 standard~\cite{DigitalIdentityGuidelines} for identity management (see OMB Memo 19-17~\cite{OMB-19-17}). NIST’s 800-63 requirements can be met using facial matching technology to enable the public to prove their identity without traveling to a government facility. Although facial matching and remote verification systems can be effective, they must be implemented carefully and co-designed with the public through thorough quantitative and qualitative research that engages a representative cross-section of users (see U.S. Office of management and Budget Memo 23-22~\cite{OMB-23-22}). Furthermore, these technologies need to be tested in real-world conditions especially given the recent learnings and concerns surrounding Artificial Intelligence models and systems (see M-24-10 on AI~\cite{OMB-AI-memo}).

By implementing an end-to-end scenario evaluation of remote identity verification (RIdV), we partnered with The United States General Services Administration (GSA) to explore whether and to what extent remote identity proofing — which the American public increasingly expects and uses — could create unforeseen barriers to accessing government services and benefits for some demographic groups. The results of this study will be a valuable reference for agencies and institutions seeking to fairly implement remote identity verification.

\section{Fairness Metrics and Statistical Methods}
There has been prior work around fairness metrics for biometric systems.  de Freitas Pereira and Marcel \cite{FDR} introduced a metric called the Fairness Discrepancy Rate (FDR). This metric provides a way to summarize how well a biometric system is performing, considering both the False Non-Match Rate (FNMR) and the False Match Rate (FMR).  Howard et al. 
 noted a scaling issue with FDR. To address this problem, they proposed a new fairness measure called the Gini Aggregation Rate for Biometric Equitability (GARBE) \cite{Demographic-Effects}.  Other metrics utilize the ratio of the performance for each category within a demographic to the minimum or geometric mean across the categories of the demographic \cite{grother2021demographic, iso_equity}. 
Other research considers the issue of detecting differences from a statistical perspective \cite{Schuckers2022}, through the use of error bounds which control the probability that differences are seen by chance. Furthermore, the research is extended to account for scenarios where each individual in a study belongs to multiple demographic groups like age, gender, and race~\cite{schuckers2023}.  This methodology is used as part of this study and is further described in the next section. 

 In this study, we focus on the false negative rate (FNR) values only (i.e.~similar to FNMR for face matching) and utilize statistical sampling variability to assess equity.  Thus, we define a RIdV technology to be \textit{inequitable} for a particular demographic (e.g.~race/ethnicity) if the FNR for one of the categories within that demographic (e.g.~Black/African American or race Asian American and Pacific Islander (AAPI)) falls outside statistical error bounds generated assuming equality of FNR across all demographics.  

\section{Data Collection}




\textbf{Participants and Demographic Targets}
Through a combination of ``on the ground/grassroots'' recruitment, social media, and survey panels, the United States General Services Administration (GSA) recruited 3,991 participants who completed the full set of checks. These participants were distributed across 5 ethnic and racial demographics: Asian American and Pacific Islander (AAPI), American Indian (AI), Black/African American (B/AA), Hispanic/LatinX (H/L), and White (Wh). There were additional participants who completed a partial set of tests but abandoned the study midway through for either technical or personal reasons but were not included in the results presented here.

\begin{figure}[!t]
\centering
\begin{minipage}[htbp]{0.45\textwidth}
    \includegraphics[width=\textwidth]{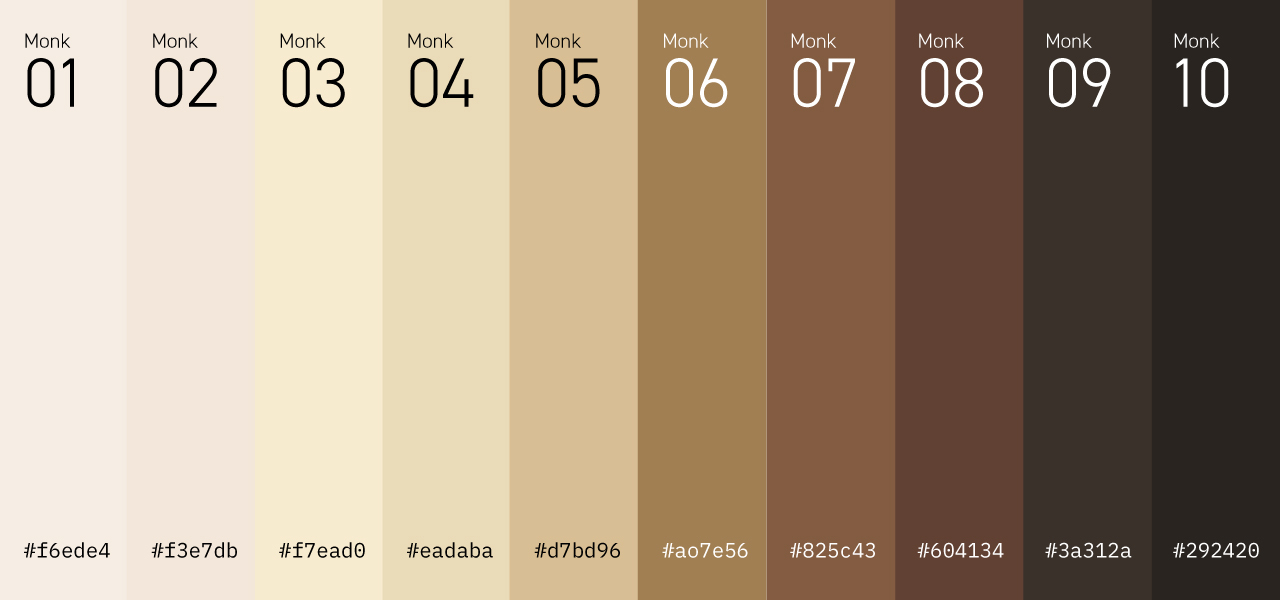}  
    \label{fig:monk scale}
    \end{minipage}
\hfill

\caption{ The Monk scale includes 10 color shades to describe human skin color \cite{Skin-Lightening}}
\label{monkscale}

\end{figure}

Participants also provided other socioeconomic data such as age, education, gender, and household income. Given recent findings on the challenge of facial matching algorithms’ performance for darker skin tones, participants were also asked to self-report their skin tone as referenced by the Monk Skin Tone scale in Figure \ref{monkscale}.  The Monk Skin Tone Scale is a simple way to describe skin colors and has 10 skin tones. For the analysis, we combined into four groups from lightest to darkest: Monk1/2/3, Monk4/5/6, and Monk7/8/9/10. The agency also recruited participants from across the U.S. and its territories to ensure that a majority of the State and Territory-issued Identification cards were tested. 

\begin{figure}[!t]
\centering


\begin{minipage}[htbp]{0.5\textwidth}
    \includegraphics[width=\textwidth]    {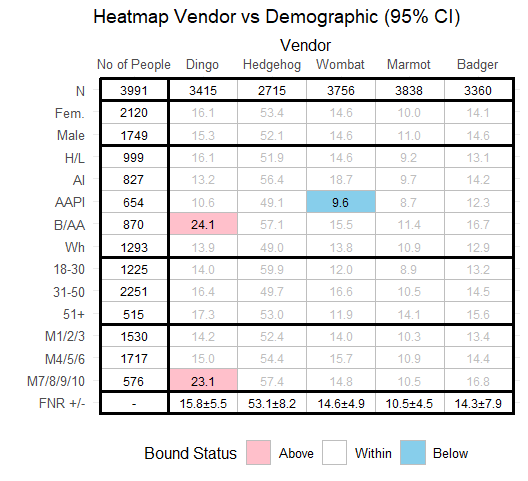}
    \label{fig:race}
    \end{minipage}
\hfill

\caption{False Negative Rates (FNR) for each demographic for each vendor.  Error rates that are outside of the 95\% confidence bounds are highlighted in pink if they are above the bounds and in blue if they are below the bounds. The number of subjects (N) for each vendor is provided in the first row and the number of subjects for each demographic group is provided in the first column. The difference in N was due to  removal of some subjects from vendors.}
\label{broadAudience2}

\end{figure}

Recruitment started in August 2023 and ended April 2024.  A total of 3,991 subjects completed the evaluation as of the date of this submission.  The number of subjects for each of the demographic groups are provided in the first column in Figure \ref{broadAudience2}.

\textbf{Vendor Selection and Algorithms}
The agency procured 11 different identity-proofing products that span three common identity checks: document authentication, facial matching, and information/identity verification. Vendors were selected from industry responses to a public ``Request for Information'' posted by the agency. Five vendors were selected due to timelines and compatibility with the study's technical platform.  The vendors are anonymized and results are listed as:  Dingo, Hedgehog, Wombat, Marmot, and Badger. 


The vendors providing document authentication and facial matching rely on different algorithms to perform these tasks. For facial matching specifically, three of the five products use algorithms tested in NIST's FRTE program (Face Recognition Technology Evaluation) while the remaining two use proprietary algorithms not yet tested in FRTE though the vendors did provide ``analogous'' report cards.

\textbf{Data Collection Procedure}
All subjects signed a consent form approved by an Institutional Review Board (IRB). Subjects first completed a survey and self-reported their demographic data.   Participants were asked to capture pictures of their identification card and a self-portrait or ``selfie” using their mobile device. Participants repeated this image capture process for each of the five randomly-ordered vendors. Participants are not made aware of RIdV software successes or failures. Vendor branding was removed on study screens to the greatest extent possible. In the background, the testing platform also collects camera resolution and other mobile device information including operating system, IP address, and web browser version.


\textbf{Privacy, Security, and Policy Compliance}
To protect personally identifiable information (PII), the study platform and the vendor connectors completed  the agency's various legal, security, and privacy processes.

All data collected during the study is kept on a Google Drive owned by the agency. RIdV providers are required to delete all data from their systems within 24 hours of each transaction. De-identified data is shared with the university collaborators for analysis.

\begin{table*}[t]
\centering
\caption{Margin for the 95\% Error Bounds for False Negative Rate for each vendor calculated using the bootstrap method }
\label{bootstrap}
\begin{tabular}{|l||c||c||c||c||c|}
\hline
& \multicolumn{1}{c||}{Dingo} & \multicolumn{1}{c||}{Hedgehog} & \multicolumn{1}{c||}{Wombat} & \multicolumn{1}{c||}{Badger} & \multicolumn{1}{c|}{Marmot} \\
\hline
95\% & 0.0547 & 0.0821 & 0.0493 & 0.0788 & 0.0448\\
\hline
\end{tabular}
\end{table*}

\section{Statistical Methodology}

In this section, we describe the statistical methodology that we used to assess whether or not a particular RIdV was statistically equitable.  Our definition of statistically equitable for a given demographic is that the false negative rate (FNR) for each category of that demographic is within a statistical error bound of that rate for all demographics.  
Our approach follows \cite{schuckers2023} 
 who controls the probability of declaring a solution inequitable when considering multiple demographics each with multiple categories based upon sample observed data when the population is equitable.  
 In the statistics literature, this probability is sometimes called the family-wise error rate or the problem of multiplicity. See, for example, \cite{kutner}. Our analysis uses bootstrapping to build a reference distribution of differences.~\cite{schuckers2023}.  For what follows we will use a 95\% confidence level and we will use a single margin across all demographic groups for a given RIdV.  This study is focused on the false negative rate, where a negative is the rejection by a vendor of a test subject. Rejections may be due to one or more sub-components, including image capture, quality check, document verification, face matching, and/or liveness (as shown in Figure \ref{rIDV_process}).

Denote the number of demographics by $D$ and let $G_d$ be the number of categories within
each demographic $d$ where $d = 1, \ldots, D$ and $k = 1, \ldots, G_d$.  Let $\hat{\pi}$ represent the estimated FNR from our sample.  The estimated FNR for category $k$ within demographic $d$ will be denoted by $\hat{\pi}_{dk}$. This is calculated by the total number of false negatives divided by the total number of attempts of individuals in that category.  
For this task, we use the methodology for a single bound, $M$, across all of the demographics for each RIdV.  
\begin{enumerate}[topsep=0pt,itemsep=-1ex,partopsep=1ex,parsep=1ex]
\item Calculate the error rate, $\hat{\pi}$ and 
the  error rate 
in each category $k$ within demographic $d$, $\hat{\pi}_{dk}$. 

\item Sample with replacement the $N$ individuals.  For the analysis below, carry along the corresponding  demographic
information (to which categories they belong) and the corresponding matching performance information (how many errors from how many attempts) for the selected individuals.

\item Calculate the bootstrapped category error rates.  Denote them as $\hat{\pi}^b_{dk}$ for each category $k$ in each demographic $d$.

\item Next calculate and store $\phi = \max_{dk} \vert \hat{\pi}^b_{dk} - \hat{\pi}_{dk} \vert$ which is the maximal difference across all demographic categories.

\item Repeat the previous three steps some large number of times, say B times.

\item Let $M$ be the $1-\alpha/2^{th}$ percentile of the distribution of $\phi$.

\item Having obtained values for $M$,  we  create a single set of 
bounds for all $\pi_{dk}$  using $\hat{\pi} \pm M$.
\end{enumerate}
We complete the algorithm above for each RIdV with $B=1000$ and  using $\alpha = 0.05$ so that we obtain 95\% error bounds for each RIdV
that are $\hat{\pi} \pm M$.  

Table \ref{bootstrap} provides the M values for each vendor for a 95\% two-sided error bound.  For example, in Table \ref{bootstrap} for System1:Dingo, $M$, and an $95\%$ confidence rate would give a range of $\hat{\pi} \pm M = 15.8\%\pm 5.5\% = (21.3\%$, $10.3\%)$.  We note that the value of $M$ is determined by a variety of factors but the most prominent is the sample size within a subgroup.  

Based upon this methodology, any demographic category falling outside the error bounds  ($\hat{\pi} \pm M$) exhibits exceptional statistical divergence from the overall False Negative Rate (FNR) will be called \textit{inequitable}.

\begin{figure*}[!t]
\centering

\title{False negative rate (FNR) with error bounds for each demographic and vendor \\
\footnotesize{Colored symbols indicate that a demographic category has FNR outside of the error bounds}\\~}


\begin{subfigure}[htbp]{0.45\textwidth}
    \includegraphics[width=\textwidth]{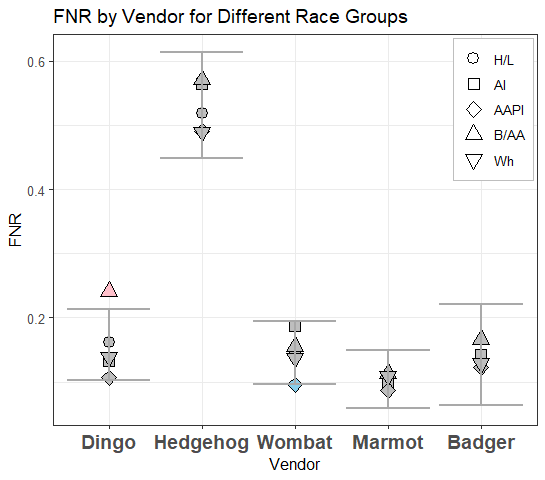}
            \caption{Racial/Ethnic Groups \label{fig:race} }
    \end{subfigure}
 \begin{subfigure}[htbp]{0.45\textwidth}
    \includegraphics[width=\textwidth]{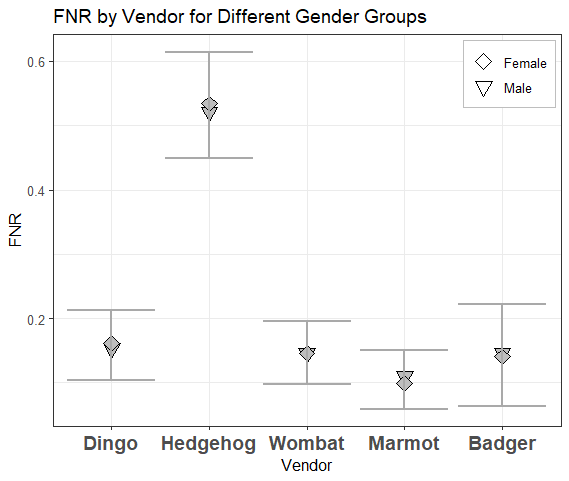}
        \caption{Gender Groups\label{fig:gender} }
    \end{subfigure}
\hfill

\vspace{0.2in}
\begin{subfigure}[htbp]{0.45\textwidth}
    \includegraphics[width=\textwidth]{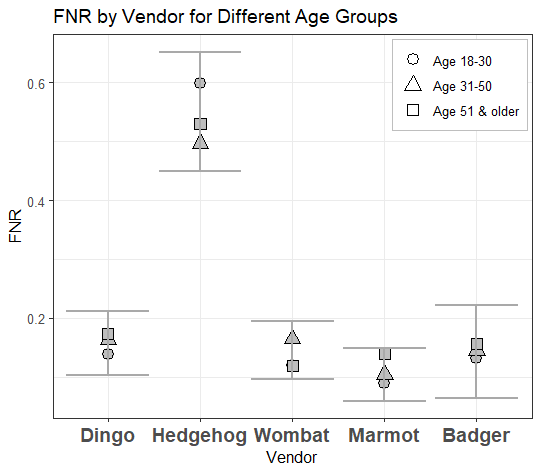}
        \caption{Age Groups\label{fig:age} }
\end{subfigure}
\begin{subfigure}[htbp]{0.45\textwidth}
    \includegraphics[width=\textwidth]{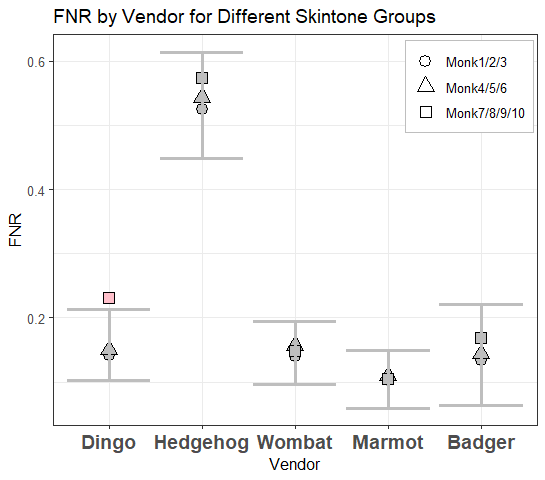}
    \caption{Monk Skin Tone Groups\label{fig:skintone} }
\end{subfigure}

\vspace{0.2in}
\caption{False negative rate (FNR) for vendors for race/ethnicity
(\ref{fig:race}), gender (\ref{fig:gender}), age (\ref{fig:age}) and skin tone based on the Monk scale (\ref{fig:skintone}).  The upper and lower bounds for the  error bounds (95\% confidence interval) are indicated by dark gray bars.  Blue and pink symbols indicate demographic categories that are below and above the bound, respectively.}
\label{fig:errorbounds}
\end{figure*}

\section{Results}
This study focuses assessing equity of Remote Identity Verification (RIdV) across age, gender, race/ethnicity, and skin tone.  
The research methodology is based on statistical approaches 
to determine whether variability among categories within each demographic group are statistically discernable.
A diverse group of 3,991 individuals utilized RIdV solutions from five vendors.  The subjects self-reported their race/ethnicity, gender, age, and skin tone.  Race had five categories:  Hispanic/LatinX (H/L), American Indian (AI), Asian American and Pacific Islander (AAPI), Black/African American (B/AA) and White (Wh).  Gender was categorized into two groups: female and male. Subjects could also report ``Other'' for gender, but those results were not analyzed due to low subject count. Age was segmented into three categories, encompassing adults aged 18-30, aged 31-50, and aged 51 and above. Finally, skin tone was reported using the Monk Scale and combined into three groups: Monk1/2/3, Monk4/5/6, and Monk7/8/9/10. 

A statistical bound (M) was computed using the bootstrap method for a 95\% error bounds for the False Negative Rate (FNR) for each vendor.   Table \ref{bootstrap} presents the margin values (M) obtained from the bootstrap percentiles for each vendor for a two-sided 95\% confidence interval.  This bound is added and subtracted to/from the mean for each vendor to produce the upper and lower bounds as shown in Figure \ref{fig:errorbounds}.

Figure \ref{broadAudience2} shows the FNR for each vendor and demographic.  The average FNR for each vendor and +/- M value for the 95\% confidence interval are provided in the last row. The boxes that are colored pink or blue are the FNR which fall outside the 95\% error bounds. For example, Marmot has a mean of 10.5\% and a margin of 4.5\%, creating a lower and an upper bound of (6\%, 15\%). The mean for Dingo is 15.8\% with a margin of 5.5\%, resulting in a lower and an upper bound of (10.3\%, 21.3\%).  Wombat shows a mean of 14.6\%, with a margin of 4.9\%, yielding a lower and an upper bound of (9.7\%, 19.5\%). Badger presents a mean of 14.3\% and a margin of 7.9\%,  resulting in a lower and an upper bound of (6.4\%, 22.2\%).   Finally, Hedgehog has a mean of 53.1\%, with a margin of 8.2\%, leading to a lower and an upper bound of (44.9\%, 63.1\%). It is observed that for the vendor Wombat, the AAPI demographic falls below the lower bound. Conversely, for the Dingo vendor, the Black/African American (B/AA) demographic exceeds the upper bound. Similarly, the Monk7/8/9/10 demographic exceeds the upper bound for the same vendor.
Figure \ref{fig:errorbounds} displays the False Negative Rate (FNR) for the five vendors for race/ethnicity, gender, age, and skin tones using the Monk scale and the upper and lower bounds for acceptable FNR values. Any demographic group exceeding the upper bound is highlighted in pink to indicate a higher FNR, while those falling below the lower bound are shown in blue, signifying a lower FNR. Groups that remain within these bounds are colored gray, denoting a case of being fair and equitable. Among the vendors, Marmot stands out for maintaining fairness and equity as all demographic groups fall within the lower and upper bounds from 6\% to 15\% and also has a notably lower FNR of 10.5\%.  The figure also illustrates that the high FNR of 53.1\% that characterizes the poor results for Hedgehog.

\section{Discussion}


The analysis of Figure \ref{broadAudience2} and Figure \ref{fig:errorbounds} reveals that Marmot exhibits consistently fair performance across all demographic groups. In contrast, Hedgehog demonstrates significantly higher error, indicating less reliable performance. Other vendors such as Wombat show specific areas of concern; notably, showing more favorable performance for the demographic group Asian American and Pacific Islander (AAPI), with the performance falling below the lower error bound. Vendors which have FNR which exceeds the upper bound are Dingo's for the race group of Black/African American (B/AA) and for Monk7/8/9/10 skin tone group. Summarizing the overall performance, Marmot leads with the most consistent and fair results across the demographics. Hedgehog, due to its high error ranges, is considered as the least reliable vendor. Between them, Wombat outperforms Dingo despite both having disparities in at least one demographic areas, with dingo suffering from higher error thresholds. 


The strategies for determining statistical differences in False Negative Rates (FNRs) across various demographic categories are based on bootstrapping \cite{Schuckers2022}. Developing robust statistical methods is essential to assess if differences in categories across various demographic groups are genuine disparities or arising by chance. Statistical boundaries, or error bounds, were calculated based on a 95\% confidence interval, and groups falling within such boundaries are considered fair, while those exceeding them  as unfair. 

One limitation of the research relates to the number of individuals for each demographic group.  
Given this study is recruiting subjects with a primary focus on race/ethnicity, other demographic categories have N values which are uneven.  
As shown in Schuckers et al.~\cite{biosig2023}, a lower value of $N$ for a specific demographic leads to an increase in the variability, which in turn causes the bounds to be wider. 
The $N$ value for each demographic category is provided in the second column of Figure \ref{broadAudience2}. 
For example, the category of Monk7/8/9/10  comprises 576 individuals and is a relatively smaller cohort compared to others, which increases the margin for all categories in the skin tone demographic (see Figure~\ref{fig:skintone}).  As such, we note that, while not significant, several other vendors (Hedgehog, Wombat, Badger) had FNR that was close to the upper bound for Monk7/8/9/10.
In future work, we will consider approaches that balance the number of subjects across demographics, such as different splits for the Monk scale or consideration of fewer demographic groups. 

Other future work will include an analysis of the causes behind various types of errors, such as failure to capture the document, document format errors (e.g. expired license), document fraud, failure to capture the face, liveness failures, and face false non-matches.

Moreover, the results only include people who completed all five vendors  without considering those who dropped out early. This could bias the results towards people who stayed until the end and may be an underestimate of the errors for a specific vendor as subjects who dropped out may be less comfortable with the technology. It should be noted that subjects were not told whether they passed or not. However, it can be hypothesized that subjects that are able to complete the full study may have better performance.

The results presented here do not include performance for fraud detection (i.e. errors where fraudulent transactions are called legitimate), as this study was focused on equity for legitimate users of the system.  Each transaction was manually reviewed to minimize the probability that fraudulent transactions were included in the  results.   Additional known fraudulent transactions were also performed randomly throughout the study to ensure that the provided vendor solution  was not tuned toward legitimate users. Fraudulent transactions were not included in the results.

Overall the error rate for even the best solution was over 10\%.   While this may seem to be a high rate of error, it must be considered in the context of the alternatives for remote identity verification.  In some use cases, the alternatives are less convenient, such as in-person enrollment, or are more fraud-prone, such as methods based only on demographic information, like providing name, birth date, and social security number.  The performance of RIdV can be further improved if solutions can be checked against an authoritative source, such as DMV records, where this service is offered.

\vspace{-0.1in}
\section{Conclusion}
\vspace{-0.1in}

Remote identity verification (RIdV) software validates identity based on a photo of an identity document, like a driver's license, and a selfie.  A successful verification evaluates the legitimacy of the document, whether the selfie face photo matches the photo on the document, and whether it is taken from a ``live'' individual.  This study evaluates RIdV technologies from five vendors to determine if their performance (false negative rate) was equitable across different demographics, including gender, age, race/ethnicity, and skin tone. The study used statistical methods to ensure any differences in performance among groups were statistically discernible, i.e., were unlikely to happen by chance. The findings revealed that one solution performed better than the others, with an error rate of about 10.5\%.  This solution also was equitable across demographic groups with all groups following within a 95\% confidence interval of 10.5\% plus or minus 4.5\%.  The other technologies were equitable across most demographics with exceptions for, Asian American and Pacific Islander (AAPI) (one vendor with more favorable performance), Black/African American (B/AA) (1 vendor with less favorable performance), or those with darker skin tones (1 vendor with less favorable performance). This research highlights the importance of evaluating RIdV technologies on diverse demographic groups to determine solutions that are fair and effective.

\vspace{-0.1in}
\section{Acknowledgments}
This work was supported by grants from the US National Science Foundation CNS-1650503 and CNS-1919554, and the Center for Identification Technology Research (CITeR).

\balance

{\small
\bibliographystyle{ieee}
\bibliography{goatssheep}
}
\end{document}